\documentclass[10pt, a4paper, conference, compsocconf]{IEEEtran}

\ifCLASSINFOpdf
  \usepackage[pdftex]{graphicx}
\else
\fi

\usepackage{url}
\begin{document}
%
\title{Object Tracking in Videos: Approaches and Issues}

\author{\IEEEauthorblockN{Duc Phu CHAU\IEEEauthorrefmark{1}\IEEEauthorrefmark{2},
Francois BREMOND\IEEEauthorrefmark{1},
Monique THONNAT\IEEEauthorrefmark{1}}

\IEEEauthorblockA{\IEEEauthorrefmark{1}Stars team, INRIA, France  \\
2004 route des Lucioles, 06560 Valbonne, France \\
Email: \{Duc-Phu.Chau, Francois.Bremond, Monique.Thonnat\}@inria.fr \\
\url{http://team.inria.fr/stars}
}
\IEEEauthorblockA{\IEEEauthorrefmark{2}  Department of Technology, Phu Xuan Private University, Vietnam \\
176 Tran Phu, Hue, Vietnam}
\url{http://www.phuxuanuni.edu.vn}
}


\maketitle

\begin{abstract}
Mobile object tracking has an important role in the computer vision applications. In this paper, we use a tracked target-based taxonomy to present the object tracking algorithms. The tracked targets are divided into three categories: points of interest, appearance and silhouette of mobile objects. Advantages and limitations of the tracking approaches are also analyzed to find the future directions in the object tracking domain.

\end{abstract}

\begin{IEEEkeywords}
 Object tracking; Computer vision; Video surveillance
 \end{IEEEkeywords}

\IEEEpeerreviewmaketitle

\section{Introduction}

\begin{figure}[]
\center
\includegraphics[width=\linewidth]{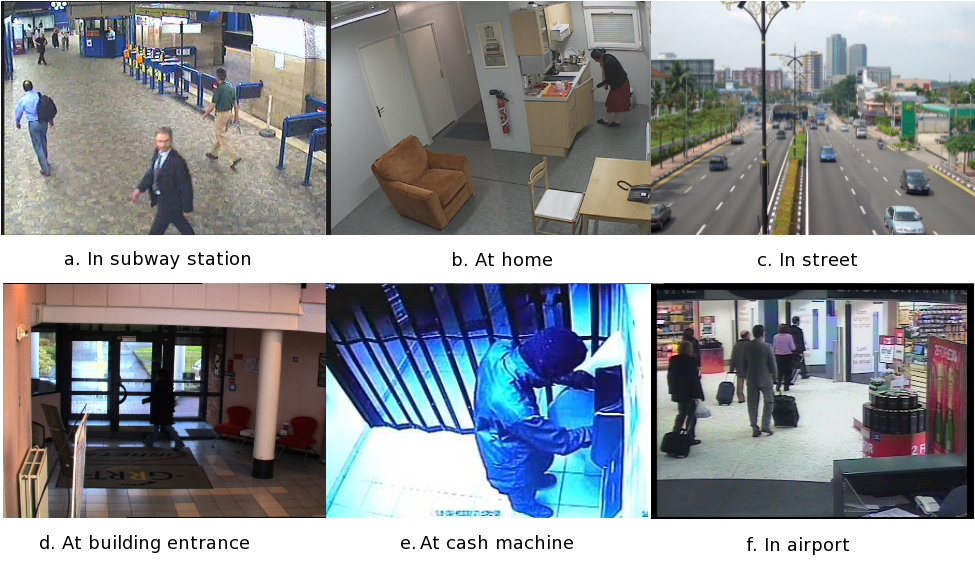}
\caption{Illustration of some areas monitored by video cameras}
\label{figCams}
\end{figure}

\begin{figure}[]
\center
\includegraphics[width=0.8\linewidth]{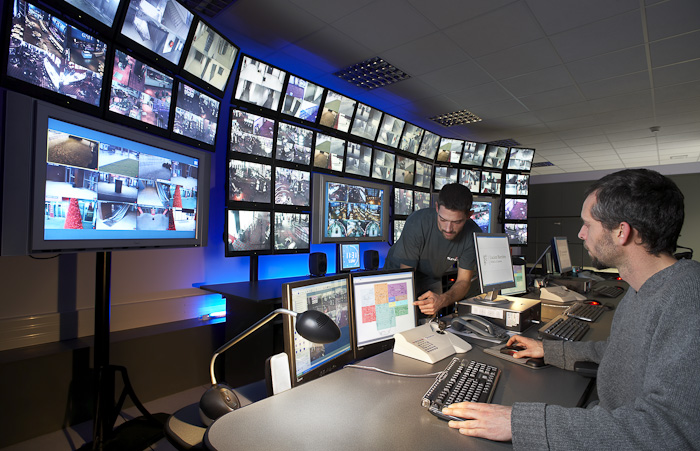}
\caption{A control room for video surveillance (source \cite{controlRoom})}
\label{figRoom}
\end{figure}

Nowadays video surveillance systems are installed
worldwide in many different sites such as airports, hospitals, banks, railway stations and even at home (see figure  \ref{figCams}). The surveillance cameras help a supervisor to oversee many different areas from the same room and to quickly focus on abnormal events taking place in the controlled space. However one question
arises: how can a security officer analyse and simultaneously dozens of monitors with a
minimum rate of missing abnormal events (see figure \ref{figRoom}) in real
time?
Moreover, the observation of many screens for a long
period of time becomes tedious and draws the
supervisor's attention away from the events of
interest. The solution to this issue lies in three
words: intelligent video monitoring.

\begin{figure}[]
\center
\includegraphics[width=\linewidth]{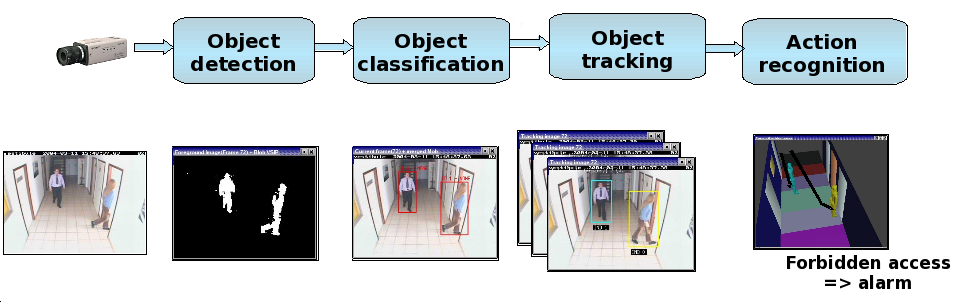}
\caption{Illustration of a video interpretation system. The first row presents the task names. The second row presents result illustrations of corresponding tasks.}
\label{figChain}
\end{figure}

The term ``intelligent video monitoring''
expresses a fairly large research direction that is
applied in different fields: for example in robotics
and home-care. In particular, a lot of researches and
works are already achieved in video surveillance applications. Figure \ref{figChain} presents a processing chain of a video interpretation system for action recognition. Such a chain includes generally different tasks: image acquisition, object detection, object classification, object tracking and activity recognition. This paper studies the mobile object tracking task.

The aim of an object tracking algorithm is to generate the trajectories of objects over time by locating their positions in every frame of video. An object tracker may also provide the complete region in the image that is occupied by the object at every time instant. Mobile object tracking has an important role in the computer vision applications such as home care, sport scene analysis and video surveillance-based security systems (e.g. in bank, parking, airport). In term of vision tasks, the object tracking task provides object trajectories for several tasks such as activity recognition, learning of interest zones or paths in a scene and detection of events of interest.

In this paper, we present a classification of tracking algorithms which is based on tracked target categories \cite{yilmaz}. For each tracker category, we present some typical approaches, their advantages as well as their limitations. This paper is organized as follows. Section \ref{sec_overview} presents the overview of the presented tracker taxonomy. Sections \ref{sec_point_tracking}, \ref{sec_appearance_tracking} and \ref{sec_silouhette_tracking} present in detail each tracker category. A conclusion is presented at section \ref{sec_conclusion}.

\section{Overview of Tracking Algorithm Classification}
\label{sec_overview}

The tracking algorithms can be classified by different criteria. In \cite{motamed}, based on the techniques used for tracking, the author divides the trackers into two categories: the model-based and feature-based approaches. While a model-based approach needs the model for each tracked object (e.g. color model or contour model), the second approach uses visual features such as Histogram of Oriented Gradients (HOG) \cite{hog}, Haar \cite{haar} features to track the detected objects. In \cite{ojeda}, the tracking algorithms are classified into three approaches: appearance model-based, geometry model-based and probability-based approaches. The authors in \cite{aggarwal} divide the people tracking algorithms into two approaches: using human body parts and without using human body parts.

\begin{figure*}[]
\center
\includegraphics[width=12cm]{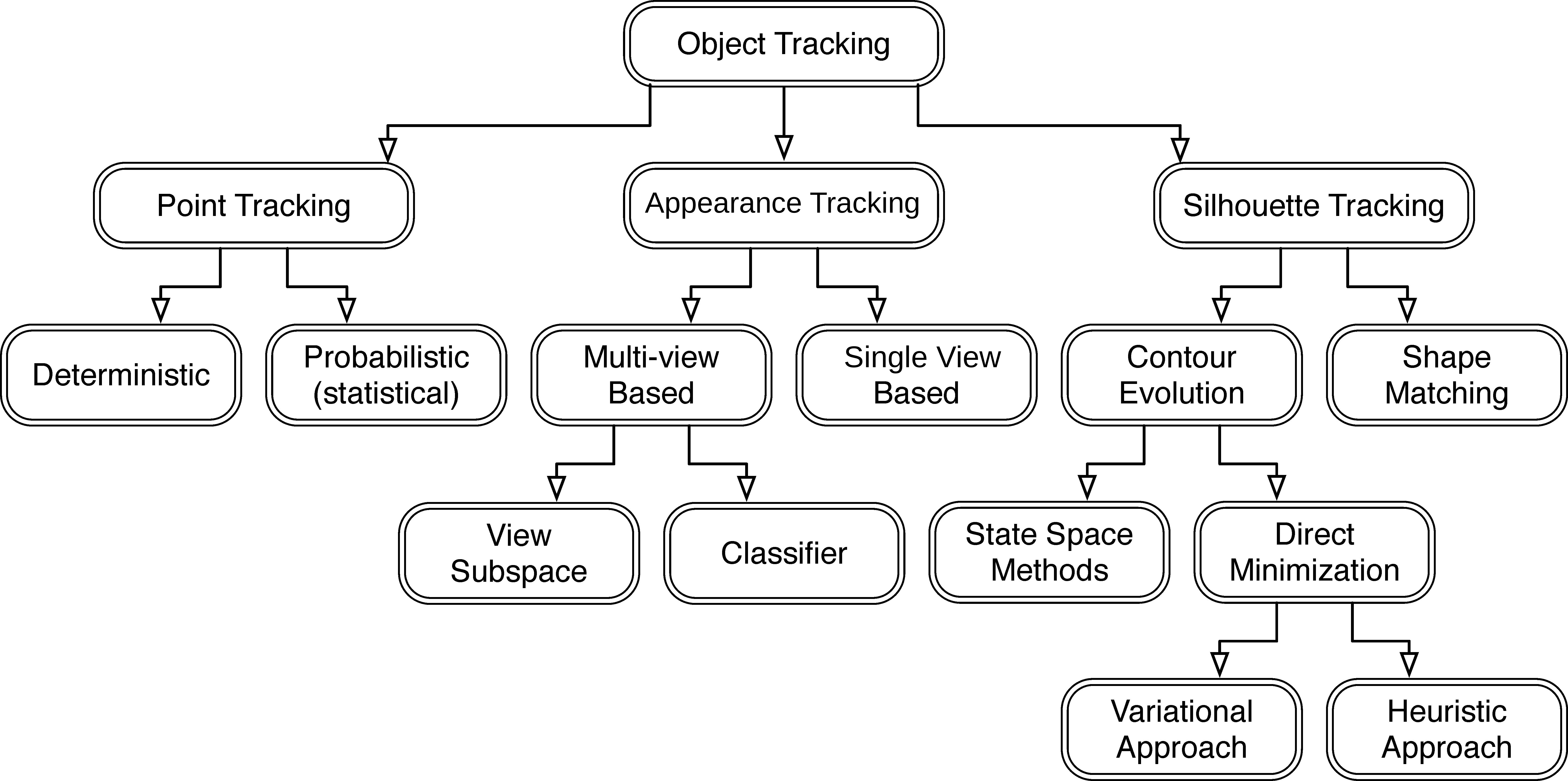}
\caption{Taxonomy of tracking methods (adapted from \cite{yilmaz}).}
\label{fig_tracker_approaches}
\end{figure*}

In this paper, we present the object tracking classification proposed by \cite{yilmaz} because this classification represents clearly and quite completely the tracking methods existing in the state of the art. This taxonomy method relies on the ``tracked targets''. The tracked targets can be points of interest, appearance or silhouette of mobile object. Corresponding to these target types, three approach categories for object tracking are determined: point tracking, appearance tracking and silhouette tracking. Figure \ref{fig_tracker_approaches} presents the taxonomy of tracking methods proposed by this paper.

\begin{figure*}
\center
\includegraphics[width=0.8\linewidth]{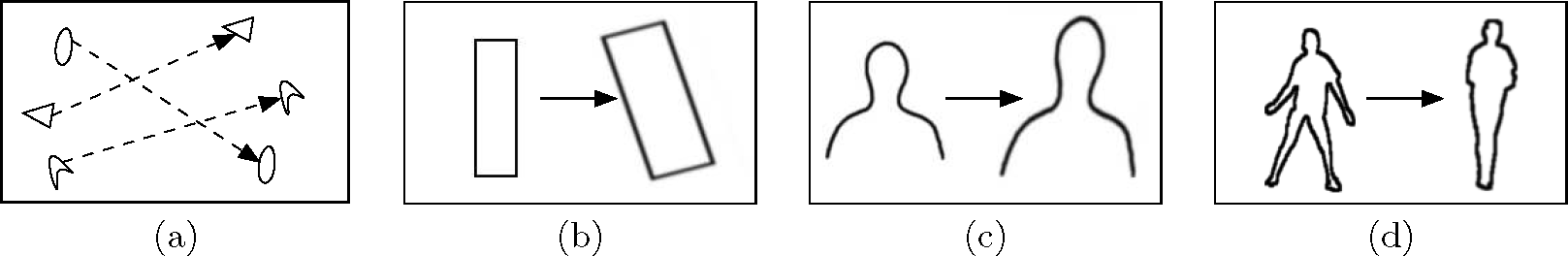}
\caption{Illustration of different tracking approaches. (a) Multipoint correspondence, (b) Parametric transformation of a rectangular patch, (c, d) Two examples of silhouette matching. (Source \cite{yilmaz}).}
\label{fig_example_trackers}
\end{figure*}

\begin{itemize}
\item \textbf{Point tracking}: The detected objects are represented by points, and the tracking of these points is based on the previous object states which can include object positions and motion. An example of object correspondence is shown in figure \ref{fig_example_trackers}(a).

\item \textbf{Appearance tracking} (called ``kernel tracking'' in \cite{yilmaz}): The object appearance can be for example a rectangular template or an elliptical shape with an associated RGB color
histogram. Objects are tracked by considering the coherence of their appearances in consecutive
frames (see example in figure \ref{fig_example_trackers}(b)). This motion is usually in the form of a parametric transformation
such as a translation, a rotation or an affinity.

\item \textbf{Silhouette tracking}: The tracking is performed by estimating the object region in each
frame. Silhouette tracking methods use the information encoded inside the object
region. This information can be in the form of appearance density and shape models
which are usually in the form of edge maps. Given the object models, silhouettes are
tracked by either shape matching or contour evolution (see figures \ref{fig_example_trackers}(c), (d)).

\end{itemize}

\section{Point Tracking}
\label{sec_point_tracking}

\subsection{Deterministic Approaches}

According to \cite{deterministic}, a deterministic system is a system in which no randomness is involved in the development of the future states of the system. A deterministic model thus always produces the same output from a given starting condition or initial state. In order to apply this idea for object tracking, the object movements are generally assumed to follow some trajectory prototypes. These prototypes can be learned offline, online, or constructed based on a scene model. We can find in the state of the art many tracking algorithms based on this idea \cite{scovanner, piotr, baiget}.

In \cite{scovanner}, the authors present a method to learn offline some tracking parameters using ground-truth data. In the offline phase, the authors define an energy function to compute the correctness of the people trajectories. This function is denoted $E(x_t)$ where $x_t$ is a $2D$ vector containing the pedestrian's location at time $t$.

The authors assume that a pedestrian path is constrained by the four following rules. Each rule is represented by an energy function.

\begin{enumerate}
\item The displacement distance of people between two consecutive frames is not too large. The energy function expressing this rule is denoted $E_{1}(x_t)$.

\item The speed and direction of people movement should be constant. The energy function corresponding to this rule is denoted $E_{2}(x_t)$.

\item People movements should reach their destinations. The energy function representing this rule is denoted $E_{3}(x_t)$.  

\item People movements intend to avoid people in the scene. The energy function of this rule is denoted $E_{4}(x_t)$.  

\end{enumerate}

\begin{figure*}[]
\center
\includegraphics[width=0.8\textwidth]{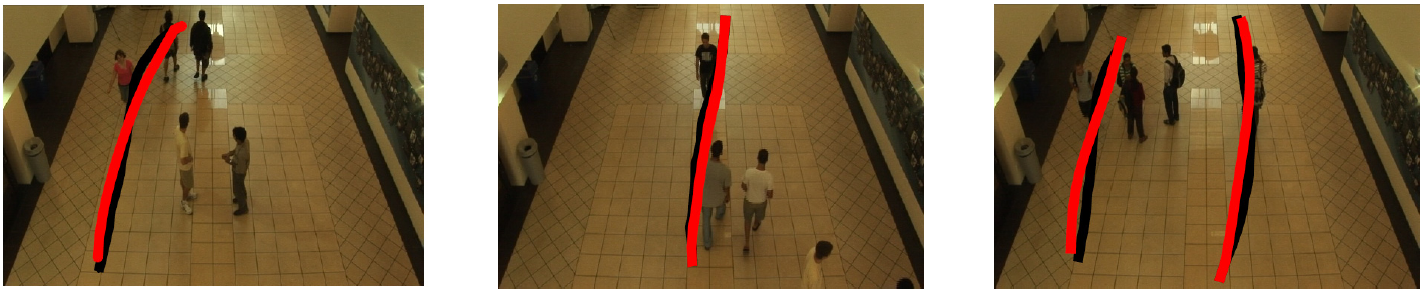}
\caption{Examples of pedestrian paths, shown in black, and predicted paths, shown in red. The model accurately predicts the deflection of pedestrians due to oncoming obstacles (source \cite{scovanner}).}
\label{fig_pedestrian_result}
\end{figure*}

The complete energy $E(x)$ is a weighted combination of these components:

\begin{equation}
E(x_t) = \sum _ {i = 1} ^ {4} \theta_i E_{i}(x_t)
\end{equation}

\noindent where $\theta_i$ represents the weight of the energy function $i$. This complete energy function is used to predict the pedestrian locations in the next frame. The pedestrians should move to the locations that minimize this energy. The objective of the training phase is to learn the values $\theta_i$ that make
the predicted pedestrian tracks match corresponding tracks in the ground-truth data. To accomplish this, the authors define
a loss function $L(x^*, g)$ that measures the difference
between a predicted track $x^*$ and the ground-truth track
$g$ as follows:

\begin{equation}
L(x^*, g) = \sum _ {i = 1} ^ {N_s} \sqrt{ || x_t - g_t ||^2 + \epsilon }
\end{equation}

\noindent where $x_t$ and $g_t$ are locations of predicted track and ground-truth track at time $t$, and $N_s$ is the number of positions of the considered track. The learned values $\theta_i$ are used later in the testing phase to predict pedestrian trajectories. Figure \ref{fig_pedestrian_result} shows some examples of the predicted paths (in red color) and their corresponding reference paths (in black color).

The advantage of this approach is that its performance does not depend on quality of the object detection process. However the used rules can be incorrect for complex people movements. The pedestrian destination can be changed. Obstacles are often not neither stable throughout the time. Pedestrian velocity is only correct if he/she is always detected correctly. Experimentation is only done with simple sequences.

\begin{figure}[]
\center
\includegraphics[width=0.8\linewidth]{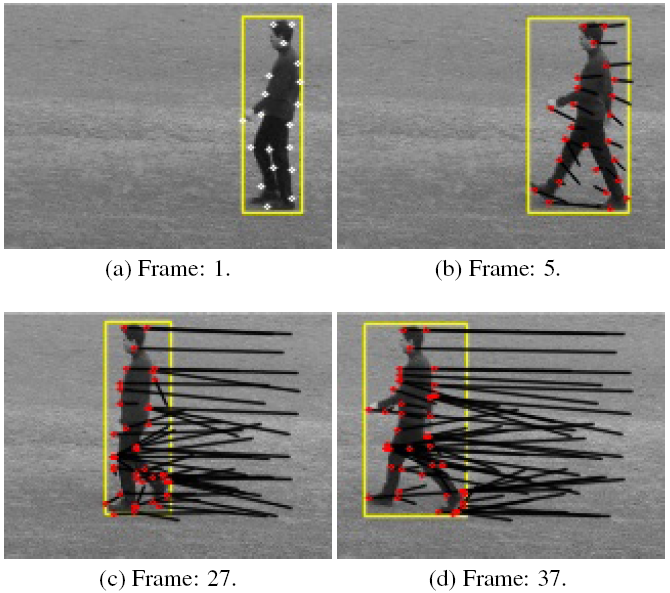}
\caption{Illustration of a point tracker for KTH dataset (source \cite{piotr}).}
\label{fig_piotr_1}
\end{figure}

\begin{figure*}[]
\center
\includegraphics[width=0.7\textwidth]{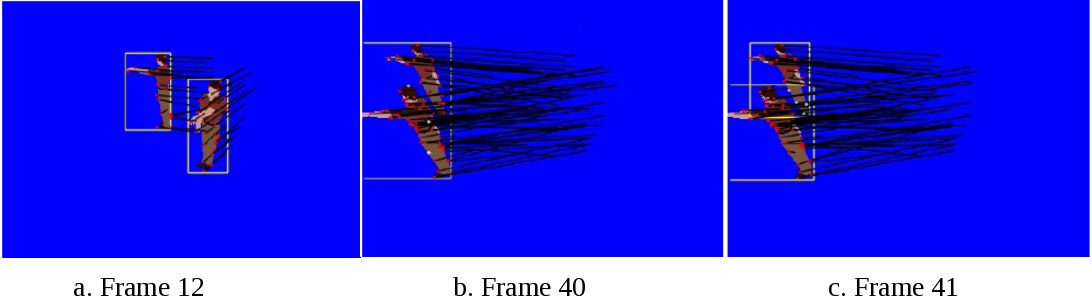}
\caption{Illustration of split a merged-object bounding box for a synthetic video sequence: a. Before merging b. Merging c. Split  (source \cite{piotr}).}
\label{fig_piotr_2}
\end{figure*}

In \cite{piotr}, the authors present a tracking algorithm based on a HOG descriptor \cite{hog}. First, the FAST algorithm \cite{fast} is used to detect the points of interest. Each point is associated with a HOG descriptor (including gradient magnitude and gradient orientation). The authors compute the similarity of the HOG points located in the consecutive frames to determine the couples of matched points. The object movements can be determined using the trajectories of their points of interest (see figure \ref{fig_piotr_1}). In the case of occlusion, the authors compare the direction, speed and displacement distance of the point trajectories of occluded objects with those of objects in previous frames to split the bounding box of occluded objects (see figure \ref{fig_piotr_2}). This approach can be well performed in the case of occlusions in which object appearance is not fully visible. However, the HOG descriptor reliability decreases significantly if the contrast between the considered object and its 
background is low.

\subsection{Probabilistic Approaches}

\begin{figure}[]
\center
\includegraphics[width=\linewidth]{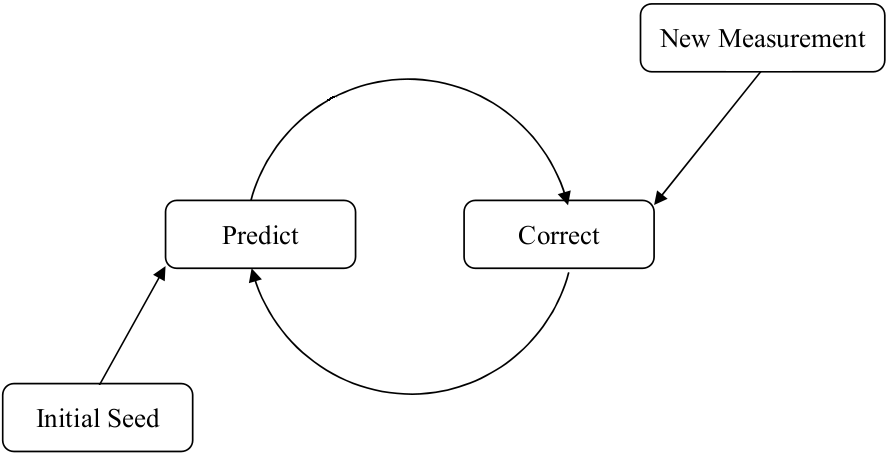}
\caption{Illustration of Kalman filter steps (source \cite{kalman_summary})}
\label{fig_kalman}
\end{figure}

Probabilistic approaches represent a set of object tracking methods which rely on the probability of object movements. In this approach, the tracked objects are represented as one or many points. One of the most popular methods of this approach is Kalman filter-based tracking. A Kalman filter is essentially a set of recursive equations that can help to model and estimate the movement of a linear dynamic system. Kalman filtering is composed of two steps: prediction and correction. The prediction step uses the state model to predict the new state of variables:

\begin{equation}
X_t^- = D X_{t-1}^+ + W
\end{equation}

\begin{equation}
{P}_t^- = D P_{t-1}^+ D^T + Q^t
\end{equation}

\noindent where $X_t^-$ and $X^+_{t-1}$ are respectively the predicted and corrected states at time $t$ and $t-1$; $P^-_t$ and $P^+_{t-1}$ are respectively the predicted and corrected covariances at time $t$ and $t-1$. $D$ is the state transition matrix which defines the relation between the state variables at time $t$ and $t - 1$, $W$ is a noise matrix, $Q$ is the covariance of the noise $W$. Similarly, the correction step uses the current observations $Z_t$ to update the object's state:

\begin{equation}
 K_t = P^-_t M^T [ M P^-_t M^T + R_t ]^{-1}
\end{equation}

\begin{equation}
X_t^+ = X^-_t + K_t [ Z_t - M X^-_t ]  \ \ \  \  \ \
\end{equation}
\
\begin{equation}
P_t = P^-_t - K_t M P^-_t \ \ \  \  \ \ \ \ \  \  \ \  \  \  \ \
\end{equation}

\noindent where $M$ is the measurement prediction matrix, $K$ is the Kalman gain and $R$ is the covariance matrix of noise in measurement. An illustration of the Kalman filter steps can be found in figure \ref{fig_kalman}. The Kalman filter is widely used in the vision community for tracking \cite{beymer, broida, brookner}.

\begin{figure}[]
\center
\includegraphics[width=0.8\linewidth]{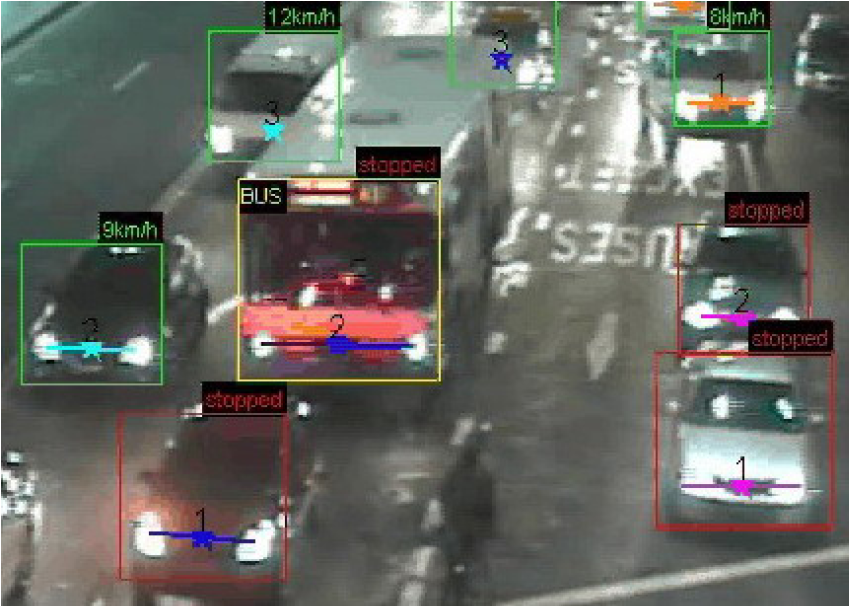}
\caption{Illustration of a night time traffic surveillance system (source \cite{robert}).}
\label{fig_night}
\end{figure}

In \cite{robert}, the authors present a tracking algorithm for vehicles during the night time (see figure \ref{fig_night}). In this work, vehicles are detected and tracked based on their headlight pairs. Assuming that the routes are linear, a Kalman filter is used to predict the movement of the headlights. When a vehicle turns, its Kalman filter is re-initialized.

\begin{figure}[]
\center
\includegraphics[width=0.8\linewidth]{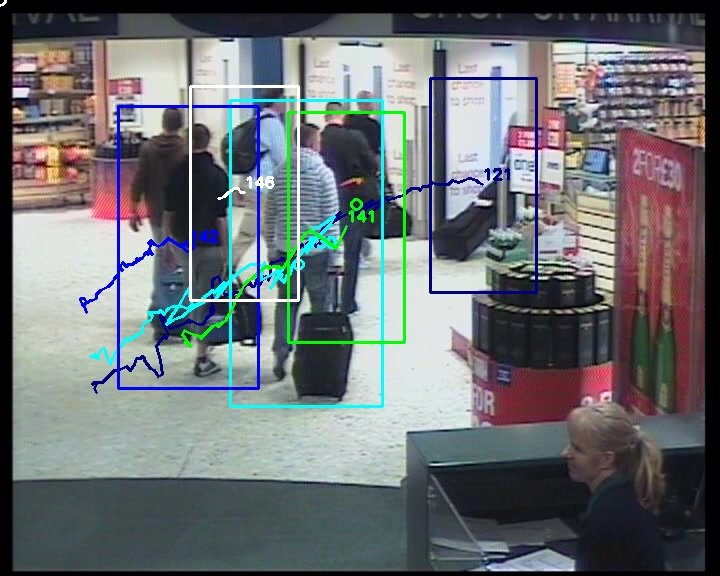}
\caption{Illustration of tracking algorithm output for TRECVid video (source \cite{dpchauVisapp11})}
\label{fig_phu_visapp11_trecvid}
\end{figure}

In \cite{dpchauVisapp11}, the authors present an algorithm to track mobile objects in different scene conditions (see illustration in figure \ref{fig_phu_visapp11_trecvid}). The main idea
of this tracker includes estimation, multi-features similarity measures and trajectory filtering. A
feature set (distance, area, shape ratio, color histogram) is defined for each tracked object to search for the best
matching object. Its best matching object and its state estimated by the Kalman filter are combined to update
position and size of the tracked object. However, the mobile object trajectories are usually fragmented because
of occlusions and misdetections. Therefore, the authors also propose a trajectory filtering, named global tracker,
aims at removing the noisy trajectories and fusing the fragmented trajectories belonging to a same mobile
object.

Because the Kalman filter assumes that the variation of the considered variables draws from a Gaussian distribution, these approaches can be only applied for tracking objects with linear movements, or with movements of simple variations of direction, speed. In order to overcome these limitations, an Extended Kalman filter \cite{extKalman} or particle filter \cite{particle} can be used.

\section{Appearance Tracking}
\label{sec_appearance_tracking}

Appearance tracking is performed by computing the motion of the object, which is represented by a primitive object region, from one frame to the next. The tracking methods belonging to this type of approaches are divided into two sub-categories: single view-based (called template-based in \cite{yilmaz}) and multi view-based.

\subsection{Single View-based Approaches}

\begin{figure}[]
\center
\includegraphics[width=\linewidth]{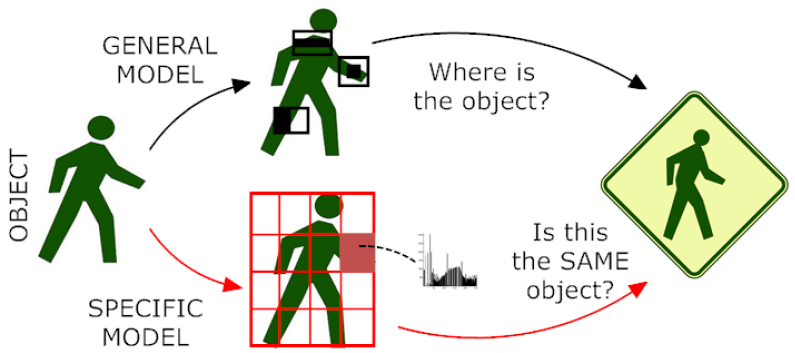}
\caption{A Haar-like features classifier is employed as a generic detector, while an online LBP features recognizer is instantiated for each detected object in order to learn its specific texture (source \cite{snidaro}).}
\label{fig_lbp_haar}
\end{figure}

\begin{figure}[]
\center
\includegraphics[width=\linewidth]{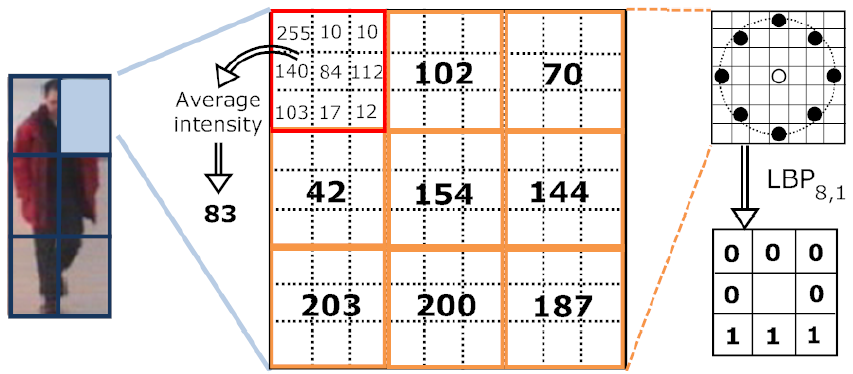}
\caption{Illustration of the LBP features computation (source \cite{snidaro})}
\label{fig_lbp}
\end{figure}

This approach category is widely studied in the state of the art for tracking mobile objects in a single camera view. Many methods have been proposed to describe the object appearance. In \cite{snidaro}, the authors present a people detection and a tracking algorithm using Haar \cite{haar} and Local Binary Pattern (LBP) \cite{lbp} features combined with an online boosting (see figure \ref{fig_lbp_haar}). The main idea is to use these features to describe the shape, the appearance and the texture of objects. While Haar features encode the generic shape of the object, LBP features capture local and small texture details, thus having more discriminative capability. First, the image is divided into cells and the Haar features are applied in each cell to detect people. Each detected person is divided into a grid of $2\ \times \ 3$ blocks. Each block is divided in 9 sub-regions. For each region, the pixel grey values are used to apply the 8-neighbours LBP calculus scheme (see figure \ref{fig_lbp}). The LBP 
features are then used to track people.

Both classifiers (Haar and LBP) are combined with an online boosting \cite{boosting}. The application of these two features in each cell (for the Haar features) or in each region (for the LBP features) is considered as the weak classifiers. These weak classifiers cluster samples by assuming a Gaussian distribution of the considered features. This online boosting scheme can help the system to adapt to specific problems which can take place during the online process (e.g. change of lighting conditions, occlusion). However, the online training is time consuming. Also, the authors do not explain clearly enough how to determine positive or negative samples in this training. It seems that the system has to learn in a sequence in which there is only one person before handling complex detection and tracking cases. The tested sequences are still simple (e.g few people in the scene, simple occlusion).

In \cite{zhou}, the authors present a method to detect occlusion and track people movements using the Bayesian decision theory. Mobile object appearance is characterized by color intensity and color histogram. For each object pair detected in two consecutive frames, if the similarity score is higher than a threshold, these two objects are considered as matched and their templates are updated. If the matching score is lower than this threshold, the authors assume that an occlusion occurs. A mobile object is divided into sub-parts and the similarity scores are computed for these parts. If the matching score of one object part is high enough while the other ones are low, an occlusion is detected. The mobile object can be still tracked but its template is not updated. This paper proposes a method to detect and track objects in occlusion cases. The authors define a mechanism to distinguish between an object appearance change due to occlusion or a real change (e.g due to the change of scene illumination or object 
distance to camera location). However the features used for characterizing the object appearance ( i.e. intensity and color histogram) are not reliable enough in the case of poor lighting condition or weak contrast. The tested video sequences are not complex enough to prove the effectiveness of this approach.

\begin{figure}[]
\center
\includegraphics[width=0.8\linewidth]{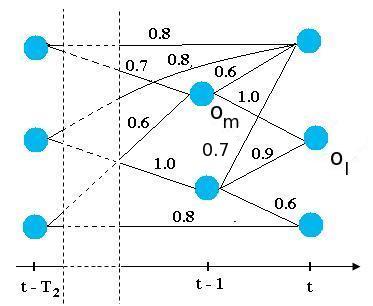}
\caption{The graph representing the established links of the detected objects in a temporal window of size T2 frames (source \cite{dpchauIcdp11})}
\label{fig_phu_Icdp11}
\end{figure}

In \cite{dpchauIcdp11}, the authors propose a tracking algorithm whose parameters can be learned offline for each tracking context. A feature pool is used
to compute the matching score between two detected objects. This feature pool includes 2D, 3D displacement distances, 2D sizes, color histogram, histogram of oriented gradient (HOG), color covariance and dominant color. An offline learning process is proposed to search for useful features and to estimate their weights for each context. In the online tracking process, a temporal window is defined to establish the links between the detected objects (see figure \ref{fig_phu_Icdp11}). This enables to find the object trajectories even if the objects are misdetected in some frames. A trajectory filter is proposed to remove noisy trajectories. However the authors suppose that the context within a video sequence is fixed over time. Moreover, the tracking context is manually selected.

\subsection{Multi-view Approach}

The methods belonging to this type of approaches are applied for tracking objects in a multi-video camera system. Because cameras can be different in rendering of colors or illumination, a color normalization step is usually necessary to make comparable object colors from different cameras (e.g. use grey intensity or compute mean and standard deviation values of color distributions).

\begin{figure*}[]
\center
\includegraphics[width=\textwidth]{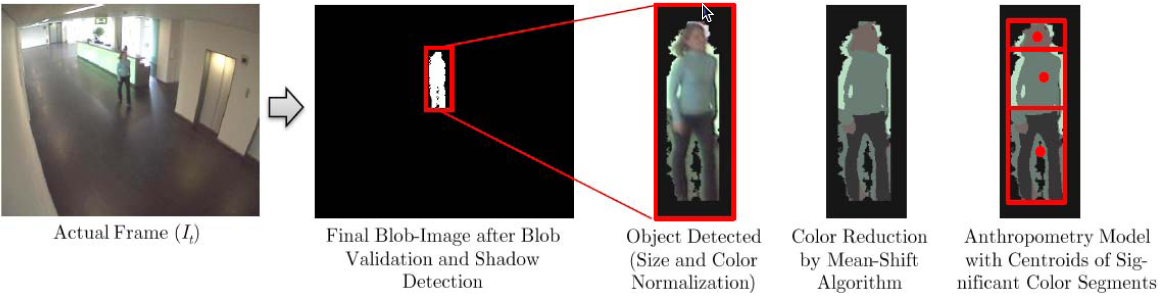}
\caption{Extraction of significant colors and mapping to dedicated body parts using a simple anthropometric model (source \cite{monari})}
\label{fig_morani}
\end{figure*}

In \cite{monari}, the authors present an appearance model to describe people in a multi-camera system. For each detected object, its color space is reduced using a mean-shift-based approach proposed in \cite{mean_shift}. Therefore, the color texture of the observed object is reduced to a small number of homogeneous colored body segments. For each color segment, the area (in pixels) and the centroid are calculated and segments smaller than 5\% of the body area are removed. Figure \ref{fig_morani} illustrates the steps of this object appearance computation. Finally, for approximative spatial description of the detected person, the person is subdivided in three sections as follows: starting from the bottom, the first 55\% as lower body, the next 30\% as upper
body, and the remaining 15\% as head. The appearance descriptor is now composed, by assigning the color segments to the corresponding body part by its centroids. Identical colors, which belong to the same body part, are merged to one. In doing so, the spatial relationships within a body part are lost but at the same time this approach leads to an invariant representation of the object in different camera views.

Let $C = (c_1, c_2,... , c_n)$ be the collection of all $n$ color segments, with $c_d = [L_d, u_d, v_d, w_d, bp_d]^T$, 

\noindent where
\begin{itemize}
 \item $d\ =\ 1..n$.
 \item $L$, $u$ are the chromatic values and $v$ is the luminance value of the homogeneous segment in CIE $L \times u \times v$ color space \cite{cie}.
 \item $w \in \{0..1\}$ (weight) is the area fraction of the color segment relative to the object area. 
 \item  $bp\ =\ \{head,\ upper body,\ lower body\}$ is the body part index which the centroid of the segment belongs to.
\end{itemize}

The appearance feature set is defined by $F^{app} \subseteq C$, with $F^{app}$ is the subset of the color segments, with a body part related fraction ($w_d$) higher than a minimum weight (e.g. 10\%). For the similarity calculation of two appearance feature sets $F_{app}^1$ and $F_{app}^2$, the Earth Mover's Distance (EMD) \cite{emd} is used.

\begin{figure}[]
\center
\includegraphics[width=0.8\linewidth]{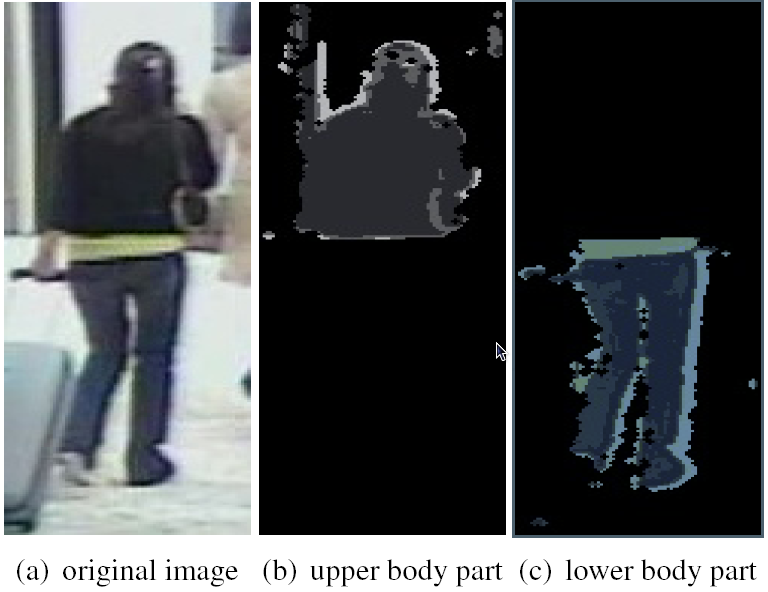}
\caption{The dominant color separation technique: a) original image; b) upper body dominant color mask; c) lower body dominant color mask (source \cite{bak_haar}).}
\label{fig_dc}
\end{figure}

In \cite{bak_haar}, the authors define a signature for identifying people over a multi-camera system. This method studies the Haar and dominant color features. For each single camera, the authors adapt the HOG-based technique used in \cite{etienne} to detect and track people. The detection algorithm extracts the histograms of gradient orientation, using a Sobel convolution kernel, in a multi-resolution framework to detect human shapes at different scales. With Haar features, the authors use Adaboost \cite{adaboost} to select the most discriminative feature set for each individual. This feature
set forms a strong classifier. The main idea of dominant color feature is to select the most significant colors to characterize the person signature. The human body is separated into two parts: the upper body part and the lower body part. The separation is obtained by maximizing the distance between the sets of dominant colors of the upper and the lower body (see figure \ref{fig_dc}). The combination of the dominant color descriptors of upper and lower body is considered as a meaningful feature to discriminate people. An Adaboost scheme is applied to find out the most discriminative appearance model.

In \cite{colombo}, the authors compare and evaluate three appearance descriptors which are used for estimating the appropriate transform between each camera's color spaces. These three appearance descriptors are: (1) mean color, (2) covariance matrix of the features: color, 2D position, oriented gradients for each channel and (3) MPEG-7 Dominant Color descriptor. In order to compare the color descriptors from two cameras, two techniques are presented to normalize color space and to improve color constancy. The first one (First-Order Normalization) consists in computing the mean value for each color component (YcbCr) over a training set of tracked objects in both cameras and to compute the linear transformation between both mean values. In the second one (Second Order Normalization), the authors consider the possibilities of rotation and translation of pixel color values. The term ``rotation'' means the difference of luminance and chromatic channels between two cameras. If there is no mixing between the 
luminance and the two chromatic channels, the rotation is not considered. The authors have tested these techniques for tracking the movement of a pedestrian over a camera network in subway stations. The result shows that the application of the color normalization techniques does not improve significantly the performance of covariance and dominant color descriptors. Also, the mean color descriptor brings the best result (compared to the two other techniques) when it is combined with the second normalization color technique. The paper gets some preliminary results on evaluation of different descriptors but the authors should extend their work on the case of multi-object tracking.

\section{Silhouette Tracking}
\label{sec_silouhette_tracking}

Objects may have complex shapes, for example, hands, head, and shoulders that cannot be well described by simple geometric shapes. Silhouette-based methods provide a more accurate shape description for these objects. The object model can be in the form of a color histogram or the object contour. According to \cite{yilmaz}, silhouette trackers are divided into two categories, namely, shape matching and contour tracking. While the shape matching methods search for the object silhouette in the current frame, the contour tracking evolves from an initial contour to its new position in the current frame by either using the state space models or direct minimization of some energy functions.

\subsection{Contour Tracking}

\begin{figure}[] 
\center
\includegraphics[width=\linewidth]{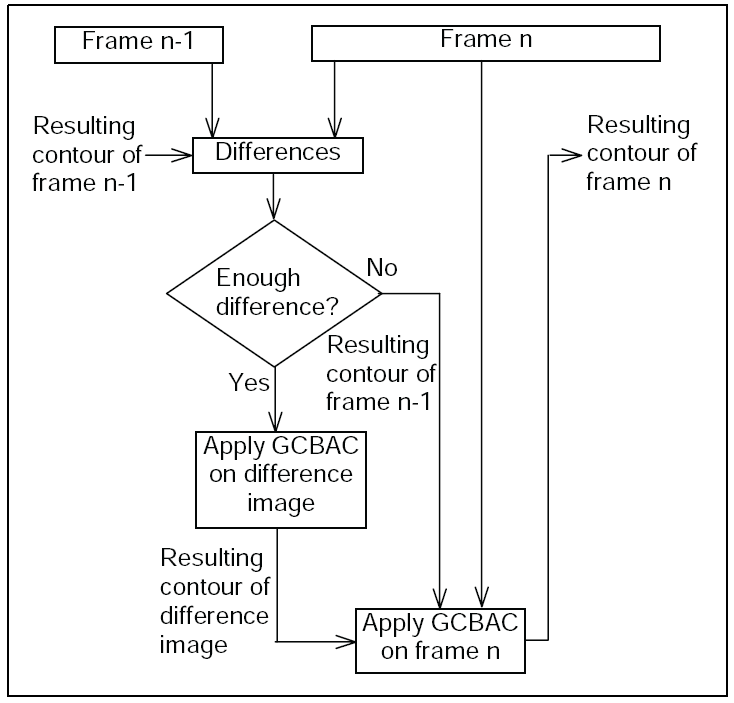}
\caption{Sketch of the object contour tracking algorithm using GCBAC (source \cite{xu02})}
\label{fig_xu02}
\end{figure}

\begin{figure}[] 
\center
\includegraphics[width=\linewidth]{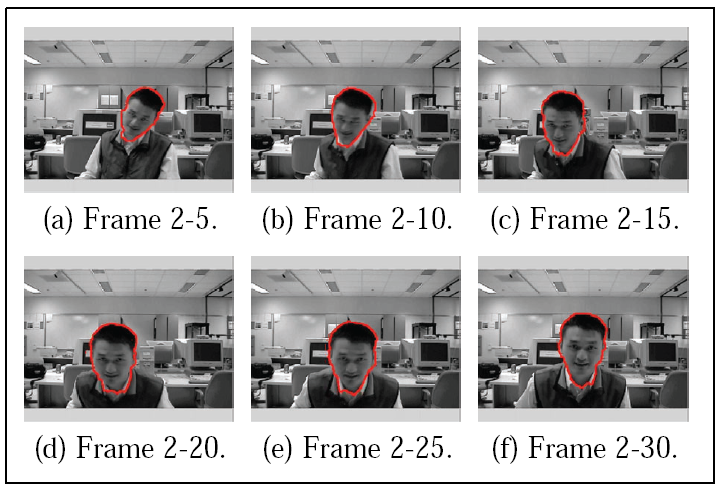}
\caption{Head tracking result when the head is rotating and translating (source \cite{xu02})}
\label{fig_xu02_result}
\end{figure}

In \cite{xu02}, the authors present an object contour tracking approach using graph cuts based active contours (GCBAC). Given an initial boundary near the object in the first frame, GCBAC can iteratively converge to an optimal object boundary. In each frame thereafter, the resulting contour in the previous frame is taken as initialization and the algorithm consists in two steps. In the first step, GCBAC is applied to the image area which is computed by the difference between a frame and its previous one to produce a candidate contour. This candidate contour is taken as initialization of the second step, which applies GCBAC to current frame directly. If the amount of difference within a neighbour area of the initial contour is less than a predefined threshold, the authors consider that the object is not moving and the initial contour is sent directly to the second step. So the initialization of the second step will be either the contour at the previous frame, or the resulting contour of the first step. Figure 
\ref{fig_xu02} presents this object contour tracking algorithm sketch. By using the information gathered from the image difference, this approach can remove the background pixels from object contour. However, this approach only works effectively if the object does not move too fast and/or the object does not change a lot in consecutive frames. It means that this approach cannot handle the cases of object occlusion. Figure \ref{fig_xu02_result} presents a head tracking result when the head is rotating and translating.

The authors in \cite{torkan} present a contour tracking algorithm based on an extended greedy snake technique combined with a Kalman filter. The contour of a mobile object includes a set of control points (called snaxels). Firstly the system computes the centroid of an object contour by calculating the average value of the coordinates of its control points. A contour is represented by its centroid and the vectors corresponding to the coordinates of the control points relatively to the centroid. The tracking algorithm then uses a Kalman filter to estimate the new centroid position in the next frame. The new control points are calculated based on this new centroid, the vectors determined in the last frame, the shape scale and the scaling factor. A new initial contour is also constructed thanks to its new control points. After that, the greedy snake technique is applied to reconstruct the contour of the mobile object. For each point of the 8 neighbour points of a snaxel, the algorithm computes a snake energy 
value and the control point is updated with the neighbour point which has minimum energy. The contour is so updated according to the new control points. The snake energy includes an internal and an external energy. In the internal energy there are continuity energy and curvature energy. While the internal energy determines the shape of the contour, the external energy prevents contour from improper shrink or shape change and always holds it close to the target boundary. In this paper, the authors use the Kalman filter to estimate the position of contour centroid in the next frame. The field energy value and the application of the Kalman filter are useful for tracking targets with high speed and large displacement. However, only three illustrations are provided. These illustrations are too simplistic. The object to track is black on a white background. A classical color segmentation should be able to detect correctly the unique object in the scene. 

\subsection{Shape Matching}

In \cite{kang}, the authors present an approach to compute the shape similarity between two detected objects. The object shape is described by a Gaussian distribution of RGB color of moving pixels and edge points. Given a detected moving blob,  a reference circle $C_R$ is defined as the smallest circle containing the blob. This circle is uniformly sampled into a set of control points $P_i$. For each control point $P_i$, a set of concentric circles of various radii are used to define the bins of the appearance model. Inside each bin, a Gaussian color model is computed for modeling the color properties of the overlapping pixels between a circle and the detected blob. Therefore, for a given control point $P_i$ we have a one-dimensional distribution $\gamma_i(P_i)$. The normalized combination of the distributions obtained from each control point $P_i$ defines the appearance model of the detected blob: $\Lambda = \sum \gamma_i (P_i) $.

\begin{figure}[] 
\center
\includegraphics[width=\linewidth]{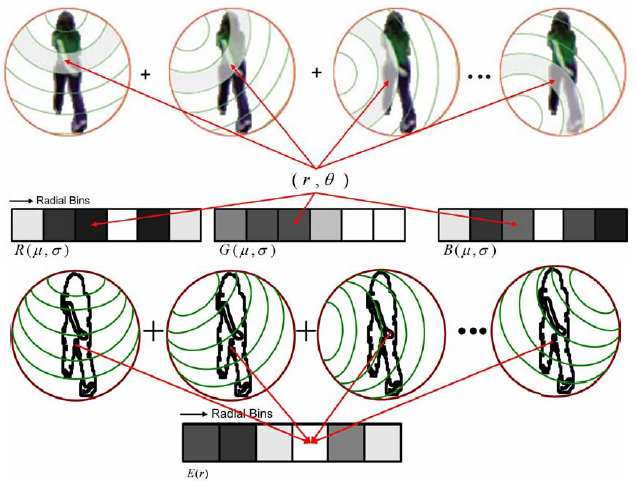}
\caption{Computation of the color and shape based appearance model of detected moving blobs (source \cite{kang})}
\label{fig_shape_matching}
\end{figure}

An illustration of the definition of the appearance model
is shown in figure \ref{fig_shape_matching} where the authors sample the reference circle with 8 control points. The 2D shape description is obtained by collecting and normalizing corresponding edge points for each bin as follows:

\begin{equation}
 E (j) = \frac{ \sum_i E_j(P_i) }{ max_j (\sum_i E_j(P_i) )}
\end{equation}

\noindent where $E(j)$ is the edge distribution for the $j^{th}$ radial bin, and $E_j(P_i)$ is the number of edge points for the $j^{th}$ radial bin defined by the $i^{th}$ control point $P_i$.

The defined model in this approach is invariant for translation. Rotation invariance is also guaranteed, since a rotation of the blob in the 2D image is equivalent to a permutation of the control points. This is achieved by taking a larger number of control points along the reference circle. Finally, the reference circle defined as the smallest circle containing the blob guarantees an invariance to scale. This approach is interesting but the authors only test with simple video sequences in which there are only two moving people.

\section{Conclusion}
\label{sec_conclusion}

This paper has presented a classification of tracking algorithms proposed in \cite{yilmaz}. The trackers are divided into three categories based on the tracked target: point tracking, appearance tracking and silhouette tracking. This classification is only relative because there are still many tracking algorithms which combine different approaches such as between point-based and appearance-based \cite{kuo}, between contour-based and shape-based \cite{yilmaz04}, or between contour-based and deterministic-based \cite{erdem03}. Understanding the advantages and limitations of each tracker category, we can select suitable tracker for each concrete video scene.

{\small
\bibliographystyle{ieee}
\bibliography{egbib}
}

\end{document}